%% file: main.tex
\documentclass[sigconf]{acmart}

\AtBeginDocument{%
  }

\usepackage[utf8]{inputenc} 
\usepackage[T1]{fontenc}    
\usepackage{hyperref}       
\usepackage{url}            
\usepackage{booktabs}       
\usepackage{amsfonts}       
\usepackage{nicefrac}       
\usepackage{microtype}      
\usepackage{xcolor}         

\usepackage{multirow}
\usepackage{algorithm}
\usepackage{wrapfig}
\usepackage{algpseudocode}
\usepackage{mathtools, amsmath}
\usepackage{bbm}
\usepackage{dsfont}
\usepackage{graphicx}
\usepackage{diagbox}

\newtheorem{problem}{Problem}
\usepackage{pifont}
\usepackage{bbding}
\definecolor{darkpastelgreen}{rgb}{0.01, 0.75, 0.24}
\newcommand{\greencheck}{{\color{darkpastelgreen}\checkmark}}
\newcommand{\redmark}{{\color{red}\ding{55}}}

\definecolor{blue}{rgb}{0.0, 0.0, 1.0}

\usepackage{makecell}
\usepackage{lipsum}

\usepackage{xspace}
\usepackage[normalem]{ulem}
\newcommand{\method}[0]{\textsc{NCPNet}\xspace}
\usepackage{subcaption}
\usepackage{balance}
\copyrightyear{2025}
\acmYear{2025}
\setcopyright{acmlicensed}
\acmConference[KDD '25]{Proceedings of the 31st ACM SIGKDD Conference on Knowledge Discovery and Data Mining V.2}{August 3--7, 2025}{Toronto, ON, Canada}
\acmBooktitle{Proceedings of the 31st ACM SIGKDD Conference on Knowledge Discovery and Data Mining V.2 (KDD '25), August 3--7, 2025, Toronto, ON, Canada}
\acmDOI{10.1145/3711896.3737064}
\acmISBN{979-8-4007-1454-2/2025/08}

\begin{document}

\title{Non-exchangeable Conformal Prediction for Temporal Graph Neural Networks}


\author{Tuo Wang}
\email{tuowang@vt.edu}
\affiliation{%
  \institution{Virginia Polytechnic Institute and State University}
  \city{Blacksburg}
  \state{VA}
  \country{USA}
}

\author{Jian Kang}
\email{jian.kang@rochester.edu}
\affiliation{%
  \institution{University of Rochester}
  \city{Rochester}
  \state{NY}
  \country{USA}
}

\author{Yujun Yan}
\email{yujun.yan@dartmouth.edu}
\affiliation{%
  \institution{Dartmouth College}
  \city{Hanover}
  \state{NH}
  \country{USA}
}

\author{Adithya Kulkarni}
\email{aditkulk@vt.edu}
\affiliation{%
  \institution{Virginia Polytechnic Institute and State University}
  \city{Blacksburg}
  \state{VA}
  \country{USA}
}

\author{Dawei Zhou}
\email{zhoud@vt.edu}
\affiliation{%
  \institution{Virginia Polytechnic Institute and State University}
  \city{Blacksburg}
  \state{VA}
  \country{USA}
}

\renewcommand{\shortauthors}{Tuo Wang, Jian Kang, Yujun Yan, Adithya Kulkarni, \& Dawei Zhou}

\newcommand{\zhou}[1]{{\small\color{red}{\bf [Zhou: #1]}}}
\newcommand{\jian}[1]{{\small\color{orange}{\bf [Jian: #1]}}}
\newcommand{\yy}[1]{{\small\color{blue}{\bf yujun: #1}}}
\newcommand{\tuo}[1]{{\small\color{purple}{\bf tuo: #1}}}

\input{0abstract.tex}

\ccsdesc[300]{Computing methodologies~Supervised learning}

\keywords{Conformal Prediction, Learning on Graphs, Temporal Graph}


\maketitle

\input{1introduction}
\input{2prelim}
\input{3approach}

\input{4algorithm}
\input{5experiments}
\input{6relatedworks}
\input{7conclusion}
\input{ack}

\bibliographystyle{ACM-Reference-Format}
\bibliography{ref}

\appendix
\input{8appendix}
\end{document}

%% file: 0abstract.tex
\begin{abstract}
Conformal prediction for graph neural networks (GNNs) offers a promising framework for quantifying uncertainty, enhancing GNN reliability in high-stakes applications. However, existing methods predominantly focus on static graphs, neglecting the evolving nature of real-world graphs. Temporal dependencies in graph structure, node attributes, and ground truth labels violate the fundamental exchangeability assumption of standard conformal prediction methods, limiting their applicability. To address these challenges, in this paper, we introduce \method, a novel end-to-end conformal prediction framework tailored for temporal graphs. Our approach extends conformal prediction to dynamic settings, mitigating statistical coverage violations induced by temporal dependencies. To achieve this, we propose a diffusion-based non-conformity score that captures both topological and temporal uncertainties within evolving networks. Additionally, we develop an efficiency-aware optimization algorithm that improves the conformal prediction process, enhancing computational efficiency and reducing coverage violations.
Extensive experiments on diverse real-world temporal graphs, including WIKI, REDDIT, DBLP, and IBM Anti-Money Laundering dataset, demonstrate \method's capability to ensure guaranteed coverage in temporal graphs, achieving up to a 31\% reduction in prediction set size on the WIKI dataset, significantly improving efficiency compared to state-of-the-art methods.
Our data and code are available at \url{https://github.com/ODYSSEYWT/NCPNET}.
\end{abstract}

%% file: 1introduction.tex
\section{Introduction}

Graph Neural Networks (GNNs) have become integral to a wide range of real-world applications, including financial fraud detection~\cite{wang2019semi}, traffic forecasting~\cite{guo2019attention}, and pharmaceutical discovery~\cite{vignac2022digress}. In high-stakes domains like these, quantifying uncertainty in model predictions is essential, as it enables human oversight when the model encounters uncertain predictions, thereby mitigating potential risks and ensuring more reliable decision-making. To achieve robust uncertainty quantification, researchers have explored various approaches, including Bayesian-based~\cite{wu2021quantifying}, Frequentist-based~\cite{kan2022multivariate}, and conformal prediction (CP) methods~\cite{vovk2005algorithmic}. Among these, CP stands out as a promising approach due to its distribution-free characteristics and ability to provide rigorous statistical guarantees on the confidence level of predictions. Unlike Bayesian or Frequentist approaches, which often rely on specific assumptions about data distributions, CP offers a flexible, theoretically grounded framework that ensures the ground truth label is included in the predicted set with a predefined level of confidence.

\begin{figure*}[ht]
    \centering
    \includegraphics[width=0.6\textwidth]{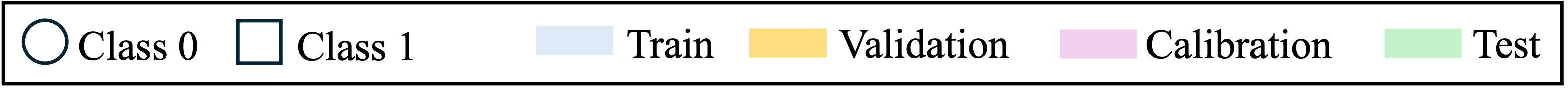}
    \begin{subfigure}[b]{0.40\textwidth}
         \includegraphics[width=\textwidth]{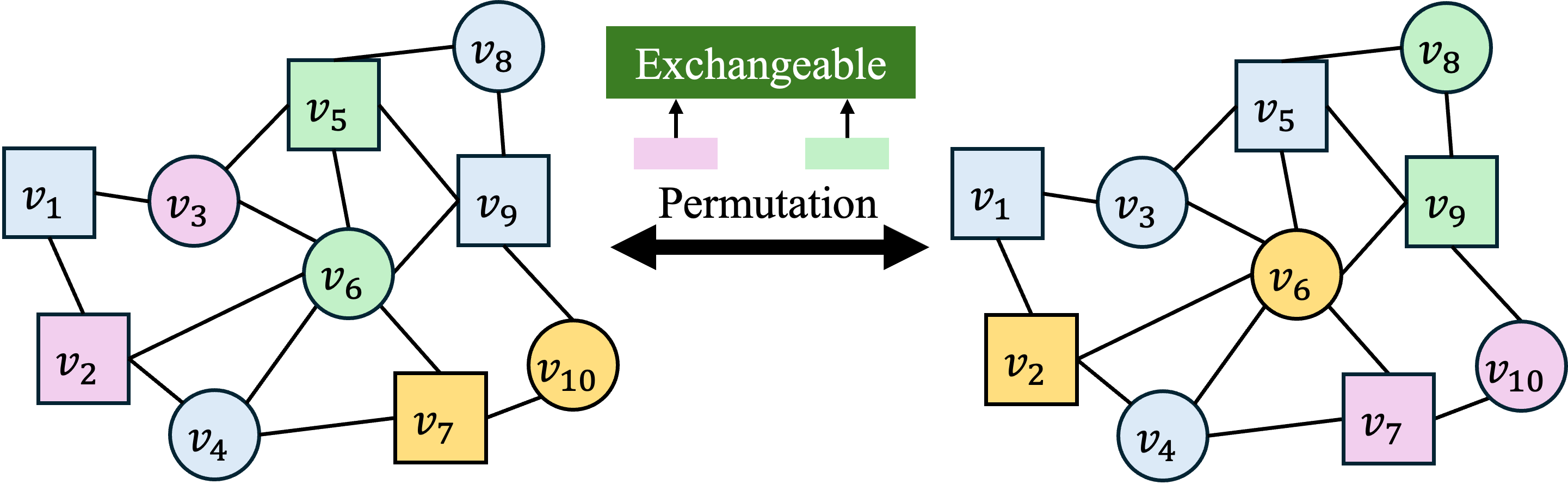}
         \caption{Static Graphs}
         \label{fig:exchangeable static graph}
    \end{subfigure}
    \hfill
    \begin{subfigure}[b]{0.59\textwidth}
         \includegraphics[width=\textwidth]{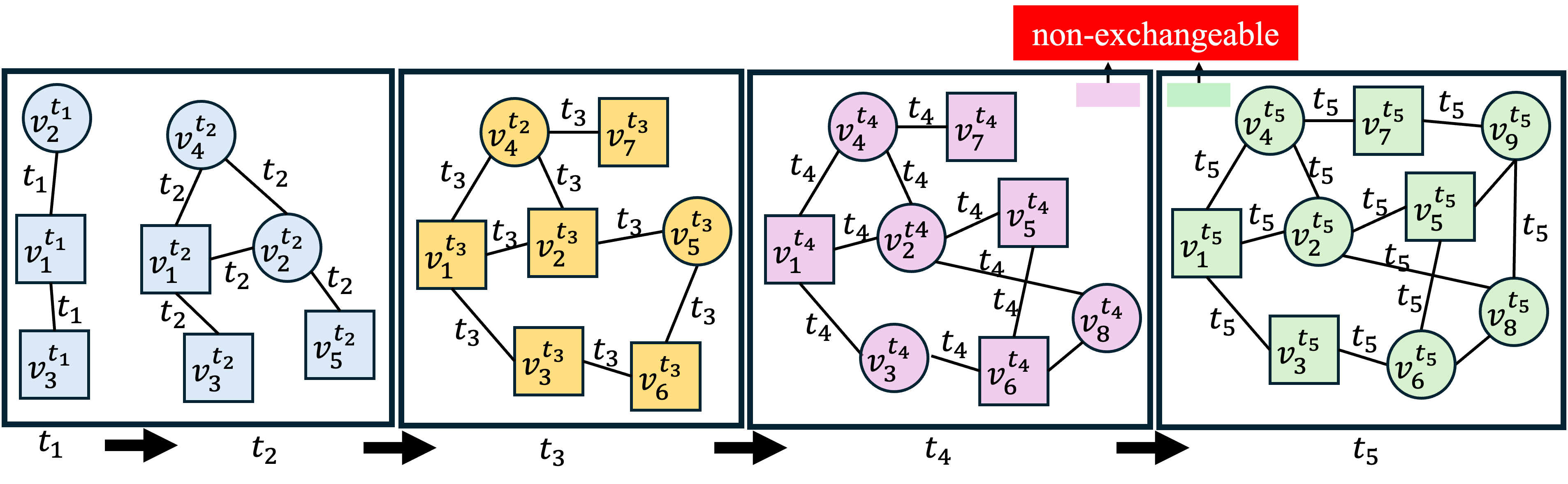}
         \caption{Temporal Graphs}
         \label{fig:non-exchangeable dynamic graph}
    \end{subfigure}
    \captionsetup{skip=2pt}
    \caption{Illustration of non-exchangeability in temporal graphs, where the shapes of nodes indicate the class memberships and the colors of nodes indicate the assignments in training set, validation set, calibration set, and test set. Figure 1 (a) shows the exchangeability between the calibration set and test set in the static graphs. Figure 1 (b) shows the non-exchangeability between the calibration set and test set on the temporal graphs. }
    \label{fig:illustration on non-exchangeable dynamic graph}
\end{figure*}

A fundamental assumption in conformal prediction (CP) is the exchangeability condition\footnote{Exchangeability definition: for any $z_1,\dots,z_{n+1}$ and any permutation $\pi$ of ${1,\dots,n+1}$, it holds that $\mathbbm{P}((Z_{\pi(1)},\dots, Z_{\pi(n+1)})=(z_{1},\dots,z_{n+1}))=\mathbbm{P}((Z_1,\dots, Z_{n+1})=(z_{1},\dots,z_{n+1}))$.}, which relaxes the independent and identically distributed (i.i.d.) assumption. This assumption generally holds in domains such as computer vision~\cite{kampffmeyer2016semantic} and natural language processing~\cite{xiao2019quantifying} because data samples are often independent of each other, making the application of CP relatively straightforward.
However, in graph-based learning, data points such as nodes and edges are inherently interconnected, leading to dependencies that violate the i.i.d. assumption and, consequently, the exchangeability condition. This violation creates significant challenges when applying CP to graphs. Recent works~\cite{pmlr-v202-h-zargarbashi23a,huang2023uncertainty} have addressed this issue for static graphs by leveraging the fact that many graph neural network architectures are permutation equivariant. This means that the structure of the graph remains unchanged under node reordering. This property allows CP to be adapted to static graphs while preserving exchangeability, as illustrated in Fig. \ref{fig:exchangeable static graph}.

Although conformal prediction has been successfully extended to static graphs, many real-world systems evolve and are represented as chronological sequences of timestamped transactions, also known as temporal edges~\cite{zhang2022deepvt}. A temporal dimension introduces fundamental challenges to the permutation equivariant properties established for static graphs, leading to a violation of exchangeability. This violation occurs because each sample in a temporal graph may follow a unique distribution influenced by temporal dependencies in graph structures, node attributes, and prediction labels. As a result, the probability of selecting different calibration sets becomes unequal, breaking the exchangeability condition. Additionally, the training process of a temporal graph inherently depends on temporal ordering, meaning that the sequence in which samples are observed directly impacts the outcomes of temporal GNNs. As illustrated in Fig. \ref{fig:non-exchangeable dynamic graph}, calibration and test sets in temporal graphs exhibit a complex relationship driven by continuity and transformation. These dependencies introduce persistent correlations and gradual shifts in the distribution of graph structures, node attributes, and ground truth labels, further complicating the application of conformal prediction.

Existing solutions for addressing non-exchangeability in temporal graphs focus on either proving exchangeability through transformations or using weighted quantile adjustments. \cite{davis2024valid} preserves exchangeability by unfolding GNNs under a stochastic block model, but this relies on a stationary stochastic process, which rarely holds in real-world temporal graphs~\cite{liao2019efficient}. In the time-series domain, \cite{barber2023conformal} proposes a non-exchangeability theory that quantifies the coverage gap and emphasizes optimized weighted quantiles to mitigate non-exchangeability. While insightful, this approach is designed for time-series data and lacks a direct method for optimizing performance in temporal graphs.

In this paper, we propose \method, a novel conformal prediction framework for temporal graphs. We begin by proving that the exchangeability condition is violated in temporal graphs, then develop a theory that quantifies the coverage gap between exchangeable and non-exchangeable settings. Our analysis shows that weighted quantiles and non-conformity measurements primarily drive this discrepancy. Building on these insights, we introduce \method, a CP algorithm designed for temporal graphs that calibrates temporal GNNs by minimizing deviations from predefined coverage. \method consists of two key components: (M1) A topological and temporal non-conformity score that improves uncertainty quantification in temporal graphs, and (M2) An efficiency-aware optimization algorithm that enhances computational efficiency and reduces the coverage gap. Our main contributions are summarized below.

\noindent $\bullet$ \textbf{Challenges in Temporal Conformal Prediction.}
We identify the challenge of non-exchangeability in applying conformal prediction to temporal graphs and formally define the conformal prediction problem in this setting. We provide theoretical proofs demonstrating that a predefined coverage level can still be guaranteed despite temporal dependencies.

\noindent $\bullet$ \textbf{Theoretical Grounding and Algorithm Design.} 
We develop a theoretical analysis quantifying the coverage gap between exchangeable and non-exchangeable conditions in temporal graphs. Our analysis reveals that weighted quantiles and non-conformity measurements influence this gap. Based on these insights, we introduce \method, a computational framework that improves conformal prediction efficiency while ensuring reliable coverage.

\noindent $\bullet$ \textbf{Empirical Evaluation.} 
We conduct extensive experiments on real-world temporal graphs to evaluate \method's effectiveness. Our results confirm that the \method consistently guarantees statistical coverage while improving efficiency, achieving up to a 31\% reduction in the prediction set size on the WIKI dataset, outperforming leading baseline methods.

%% file: 2prelim.tex
\section{Preliminary}

This section introduces the notations and the background to our problem setting. We adopt a notation convention where regular letters denote scalars (e.g., $\eta$), boldface lowercase letters represent vectors (e.g., $\mathbf{x}$), and boldface uppercase letters signify matrices (e.g., $\mathbf{X}$). A summary of key symbols can be found in Appendix \ref{ap: notation}.

\noindent \textbf{Temporal Graphs.}
The temporal graph is defined as a collection of temporal edges rather than a series of discrete snapshots~\cite{zhou2020data,fu2020local}. 
Each node is linked to multiple timestamped edges at varying times. These temporal graphs are represented as $\tilde{\mathcal{G}}=(\tilde{\mathcal{V}}, \tilde{\mathcal{X}}, \tilde{\mathcal{E}})$, where each node $v$ in $\Tilde{\mathcal{V}}$ corresponds to distinct occurrences $\{v_1^{1},...,v_n^{\mathcal{T}}\}$ along with their associated timestamped edges $\Tilde{\mathcal{E}}=\{e_{1}^{1},...,e_{m}^{\mathcal{T}}\}$, where $e_i^{t_{i}}=(v_{j}, v_{k})^{t_{i}}$. We denote the corresponding input features $\{x_{1}^{1},...,x_{n}^{\mathcal{T}}\}$ and labels $\{y_{1}^{1},...,y_{n}^{\mathcal{T}}\}$. 

\noindent \textbf{Conformal Prediction on Static Graphs.} Conformal prediction approaches are generally classified into two categories: full conformal prediction (FCP) and split conformal prediction (SCP)~\cite{vovk2005algorithmic}. FCP provides the most versatile form of CP, but the computation cost is intense since FCP needs to build a model for each calibration sample. SCP achieves a better balance between computational cost and performance. This paper focuses on SCP due to its optimal trade-off between performance and computational efficiency. We provide the theory of conformal coverage guarantee in Theorem \ref{th: split conformal prediction}, which ensures that the ground truth label is included within the prediction set with a probability of at least $1 - \alpha$.
\begin{theorem}[Conformal Coverage Guarantee~\cite{vovk2005algorithmic}] \label{th: split conformal prediction} 
Given a set of data points $(\mathbf{x}_1, y_1),\ldots,(\mathbf{x}_n, y_n),(\mathbf{x}_{n+1}, y_{n+1})$ and a desired coverage level $1 - \alpha \in (0, 1)$, a score function that maps data points from $\mathcal{X} \times \mathcal{Y}$ to $\mathcal{R}$. The prediction set is given by $C(\mathbf{x}_{n+1})=\{y: s(\mathbf{x}_{n+1}, y) \geq q\}$, where $q$ is defined as the $\frac{n-1}{n} (1-\alpha)$th smallest value from $\{s(\mathbf{x}_i, y_i): i \in (1,.., n)\}$. Thus, we can obtain the prediction set based on this scoring criterion as follows:
\begin{equation}
    \mathbbm{P}(y_{n+1} \in \mathcal{C}(\mathbf{x}_{n+1})) \geq 1-\alpha.
\end{equation}
\end{theorem}

The exchangeability assumption restricts the application of CP in graph domains, as nodes and edges exhibit dependencies that violate this condition. However, recent studies~\cite{huang2023uncertainty, pmlr-v202-h-zargarbashi23a} show that exchangeability can be preserved in node classification tasks if the non-conformity scores of a GNN are permutation invariant in static graphs. In a static graph $\mathcal{G}=(\mathcal{V}, \mathcal{E})$, each node $v \in \mathcal{V}$ has associated attributes and labels, denoted as $\mathbf{x}$ and $y$. Given training dataset $\mathcal{D}_{train}$, validation dataset $\mathcal{D}_{valid}$, calibration dataset $\mathcal{D}_{calib}$, and test dataset  $\mathcal{D}_{test}$, if a model mapping data points $\mathcal{X}$ to $\mathcal{Y}$ and a function mapping $\mathcal{X} \times \mathcal{Y}$ meet the assumption in Eq.~\ref{eq: permutations invariant}, the exchangeability condition can be maintained.
\begin{equation} \label{eq: permutations invariant}
    \begin{split}
        & S(x,y; \{x_v, y_v\}_{v\in \mathcal{D}_{\rm train} \cup \mathcal{D}_{\rm valid}}, \{\mathbf{x}_v\}_{v\in \mathcal{D}_{\rm calib} \cup \mathcal{D}_{\rm test}}, \mathcal{V}, \mathcal{E}) = \\ 
        & S(x,y; \{x_v, y_v\}_{v \in \mathcal{D}_{\rm train} \cup \mathcal{D}_{\rm valid}}, \{\mathbf{x}_{\pi(v)}\}_{v \in \mathcal{D}_{\rm calib} \cup \mathcal{D}_{\rm test}}, \mathcal{V}_\pi, \mathcal{E}_\pi),
    \end{split}
\end{equation}
where $S$ denotes the non-conformity score function, and $(\mathcal{V}_\pi,\mathcal{E}_\pi)$ represents a static graph where the nodes in $\mathcal{D}_{calib} \cup \mathcal{D}_{test}$ are permuted according to a permutation $\pi$. Typical GNN models satisfy the assumption in Eq.~\ref{eq: permutations invariant}, as they rely solely on graph structures and attributes without considering node order. However, inherent dependencies across time in temporal graphs create unequal permutation probabilities, violating this assumption and resulting in non-exchangeability. Real-world graphs evolve, requiring models to capture dynamic node relationships. These temporal dependencies further violate exchangeability, posing challenges for applying CP. Existing work~\cite{davis2024valid} preserves exchangeability under a stochastic block model, but this assumption often fails in practice. In the time-series domain, where similar issues arise, \cite{barber2023conformal} proposes a non-exchangeability theory that generalizes to both exchangeable and non-exchangeable conditions.

\noindent \textbf{Problem Definition.} With the aforementioned notations, we formally define the problem of conformal prediction for temporal graphs where the exchangeability condition is not satisfied.
\begin{problem}[Conformal Prediction in Temporal Graphs] \textbf{Given}: (i) a temporal graph $\tilde{\mathcal{G}}=(\tilde{\mathcal{V}}, \tilde{\mathcal{X}}, \tilde{\mathcal{E}})$, where $\tilde{\mathcal{V}} = \{v_1^{1},... v_n^{\mathcal{T}}\}$, $\Tilde{\mathcal{E}}=\{e_1^{1},...,e_m^{\mathcal{T}}\}$, $\tilde{\mathcal{X}}=\{\mathbf{x}_1^{1},..,\mathbf{x}_n^{\mathcal{T}} \}$, and $\tilde{\mathbf{Y}}=\{y_1^1,...,y_n^\mathcal{T} \}$ as the ground truth labels; (ii) a temporal GNN $f(\Tilde{G}, \Tilde{X}, \Tilde{\mathcal{\theta}})$; (iii) a pre-defined mis-coverage level $\alpha$. \\
    \textbf{Find}: A prediction set with the guarantee that the probability of the ground truth falls within the prediction set with a confidence level of at least $1 - \alpha$ while maintaining high efficiency.
    \label{problem_1}
\end{problem}

%% file: 3approach.tex
\section{Theoretical Analysis}
In this section, we establish the theoretical foundation for conformal prediction in temporal graphs, as formulated in Problem \ref{problem_1}. We first analyze the violation of exchangeability in temporal graphs (Proposition \ref{th: assumption not hold in temporal graph}) and then derive a theoretical bound for conformal coverage in non-exchangeable settings (Lemma \ref{th: nonexchangeble CP}).
Before presenting the theoretical framework, we formally define the calibration and test sets in the context of temporal graphs.
\begin{definition}[Calibration Set and Test Set]
    Given a set of nodes from a temporal graph denoted as $\tilde{\mathcal{V}}_{ct}=\{v_1^{t_1},\ldots,v_n^{t_n}\} \subset \tilde{\mathcal{V}}$ and the corresponding non-conformity score for $\tilde{\mathcal{V}}_{ct}$ is listed as $S_{ct}=\{s_1^{t_1},\ldots,s_n^{t_n}\}$ where $t_1 \leq \ldots \leq t_n$. 
    The calibration and test set is defined as a subset of $\tilde{\mathcal{V}}_{ct}$,
    \begin{equation}
        \mathcal{D}_{c}=\{v_i^{t_i}, \mathbf{x}_i^{t_i}, y_i^{t_i}, s_i^{t_i}\}, \tilde{\mathcal{V}}_{c}=\{v_i^{t_i}\}, i \in \{1,...,n_c\}, n_c<n,
    \end{equation}
    \begin{equation}
        \mathcal{D}_{t}=\{v_j^{t_j}, \mathbf{x}_j^{t_j}, y_j^{t_j}, s_j^{t_j}\}, \tilde{\mathcal{V}}_{t}=\{v_j^{t_j}\}, j \in \{1,...,n_d\}, n_t<n,
    \end{equation}
    where $\mathcal{D}_{c} \cap \mathcal{D}_{t} = \emptyset$, $\tilde{\mathcal{V}}{c} \cap \tilde{\mathcal{V}}_{t} = \emptyset$.
\end{definition}

Given the calibration and test set, we provide why the assumption in Eq. \ref{eq: permutations invariant} does not hold in temporal graphs. The reasons are twofold: (i) Non-conformity scores in calibration and test sets have inherent dependencies over time, making each permutation's probability unequal. We provide proof listed in Proposition \ref{th: assumption not hold in temporal graph} to demonstrate our statement. (ii) Training temporal GNNs requires temporal information, implying that node order influences GNNs' training results, thus violating the assumption in Eq. \ref{eq: permutations invariant}. As shown in Fig. \ref{fig:non-exchangeable dynamic graph}, suppose we train a temporal GNN using the graph and nodes observed at times $t_1,t_2,t_3$ and designate nodes at $t_4$ and $t_5$ as the calibration set and test set. In this scenario, we observe that node $v_8^{t_4}$ is not present in the training data, and the adjacency matrix at time $t_4$ differs significantly from those at $t_1, t_2,t_3$ and $t_5$. Hence, the temporal GNN is more likely to produce inaccurate predictions for node $v_8^{t_4}$, which affects the calibration quality and the reliability of predictions in the test set.

\begin{proposition}[Non-Exchangeability in Temporal Graphs] \label{th: assumption not hold in temporal graph}
    In the condition that there exists a $t_i$ where $(s_1^{t_1},\ldots,s_i^{t_i}) \sim P_t$ and $(s_{i+1}^{t_{i+1}},\ldots,s_n^{t_n}) \sim P_{t+\Delta_t}$, the probability of selecting the $n_c$ non-conformity scores to be included in the calibration set is represented as $P(\tilde{\mathcal{V}}_c | \tilde{\mathcal{V}}_{ct}) = \prod_{P_t} p_t \prod_{P_{t+\Delta_t}} p_{t+\Delta_t}$, such that for every permutation $P(\tilde{\mathcal{V}}_c | \tilde{\mathcal{V}}_{ct}) \neq  P(\tilde{\mathcal{V}}_{\pi(c)} | \tilde{\mathcal{V}}_{ct})$
\end{proposition}
Proposition \ref{th: assumption not hold in temporal graph} demonstrates that the probability of all the permutations of selecting $n_c$ nodes to be in the calibration set are not equal, thus compromising the exchangeability condition. The detailed proof is provided in Appendix \ref{ap: lemma proof}. 

To address the challenges of non-exchangeability in temporal graphs, we extend the existing coverage bounds for static graphs to temporal settings by relaxing the exchangeability requirement. To account for the impact of temporal dependencies, we introduce an additional compensation term that quantifies the coverage gap between exchangeable and non-exchangeable conditions. The coverage gap, which measures the difference between exchangeable and non-exchangeable conditions given a temporal graph and corresponding calibration and test set, is defined as
\begin{definition}[Coverage Gap in Temporal Graph]
    Assuming $d_{j_{test}}=(\mathbf{x}_{j_{t}}, y_{j_{t}})$ is one random selected data point from $\mathcal{D}_{t}$ and there are $n_{c}$ data points in the calibration set $\mathcal{D}_{c}$. All the data points from $\mathcal{D}_{c}$ and the one data point from $\mathcal{D}_{t}$ formalize a set with  $n_{c} + 1$ data points. Let $\mathcal{C}_{j_{t}}$ be the prediction set for the random selected test point $d_{j_{t}}=(\mathbf{x}_{j_{t}}, y_{j_{t}})$. The coverage gap is defined as:
    \begin{equation}
        \delta_{gap} = (1 - \alpha) - \mathbbm{P}\{y_{j_{t}} \in \mathcal{C}_{j_{t}} \}.
    \end{equation}
\end{definition}

Existing conformal prediction (CP) methods for graphs focus on static graphs and assume exchangeability~\cite{huang2023uncertainty, pmlr-v202-h-zargarbashi23a}. However, we demonstrate in Proposition \ref{th: assumption not hold in temporal graph} that this assumption does not hold in temporal graphs. While prior work~\cite{barber2023conformal} extends CP to non-exchangeable settings, our approach explicitly adapts this theory to temporal graphs, minimizing the coverage gap through an end-to-end optimization of weighted quantiles. Other studies~\cite{davis2024valid} attempt to prove exchangeability in temporal graphs using unfolding GNNs under a stochastic block model assumption. However, we focus on measuring and minimizing the coverage gap between exchangeable and non-exchangeable conditions. This strategy enables our approach to handle both cases effectively while ensuring higher efficiency. Furthermore, our experiments show that unfolding GNNs have a high memory cost, making them impractical for large and sparse temporal graphs.

By leveraging the definitions of the calibration set, test set, and coverage gap, we derive an upper bound for the coverage gap. This bound highlights weighted quantiles and non-conformity measurements as key factors influencing the theoretical coverage in temporal graphs, guiding the \method framework to maintain empirical coverage guarantees while optimizing efficiency. Our theory is inspired by~\cite{barber2023conformal}, but we extend it to temporal graphs, with unique challenges like non-Euclidean structure and evolving dependencies. \cite{barber2023conformal} focuses on a sequential time series, whereas we address temporal graphs with dynamic topologies. In our M2 (Section \ref{sec: framework overview}), we optimize weights for efficiency without violating theoretical assumptions, as Lemma~\ref{th: nonexchangeble CP} permits efficiency-aware weights and arbitrary non-conformity scores.
\begin{lemma}[Upper Bound for the Coverage Gap] \label{th: nonexchangeble CP}
    The coverage gap for the test data points $d_{j_{t}}=(\mathbf{x}_{j_{t}}, y_{j_{t}})$ in $\mathcal{D}_{t}$ can be bounded by
    \begin{equation} \label{eq: upper bound}
        \delta_{gap} \leq \frac{\sum_{i=1}^{n_{c}}\omega_i d_{TV}(\mathcal{\phi}, \mathcal{\phi}^i)}{1 + \sum_{i=1}^{n_{c}} \omega_i},
    \end{equation}
    where $d_{TV}$ is the total variation distance\cite{clarkson1933definitions}, the parameters $\mathbf{\omega}$ are user-defined weights such that the lower bound is likely to be small. All the non-conformity score for each data point from $\mathcal{D}_{c}$ and the selected test point $d_{j_{t}}$ forms a set called $\mathcal{\phi}$ and $\mathcal{\phi}^i$ denotes a permutation by swapping the test data point $d_{j_{t}}$ with $i$th data point in $\mathcal{D}_{c}$, which means that
    \begin{equation}
       \mathcal{\phi}=(s_1,\ldots,s_{n_{c}},s_{j_{t}}).
    \end{equation}
    \begin{equation}
        \mathcal{\phi}^i=(s_1,\ldots,s_{i-1},s_{j_{t}},s_{i+1},\ldots,s_{n_{c}},s_i).
    \end{equation}
    The coverage of the test point can be written as
    \begin{equation}
        \mathbbm{P}\{ \mathbf{y}_{j_{t}} \in \mathcal{C}_{j_{t}}\} \geq 1 - \alpha - \frac{\sum_{i=1}^{n_{c}}\omega_i d_{TV}(\mathcal{\phi}, \mathcal{\phi}^i)}{1 + \sum_{i=1}^{n_{c}} \omega_i}.
    \end{equation}
\end{lemma}

\noindent \textbf{Remark 1}: The upper bound presented in Eq. \ref{eq: upper bound} quantifies the deviation from the desired coverage when the exchangeability condition is violated. This bound remains valid regardless of whether the exchangeability assumption holds. Specifically, when exchangeability is satisfied, the total variation distance between $\mathcal{\phi}$ and $\mathcal{\phi}^i$ is zero, indicating that the data points in the validation set and the test set follow the same distribution. More generally, this upper bound provides a unified framework that encompasses both exchangeable and non-exchangeable settings in conformal prediction, making it particularly well-suited for real-world temporal graphs.

\noindent \textbf{Remark 2}: Lemma \ref{th: nonexchangeble CP} highlights the critical role of parameters $\mathbf{\omega}$ in reducing the coverage gap under non-exchangeability conditions. However, prior work \cite{barber2023conformal} does not provide a principled method for selecting these parameters, leading to inefficiencies in evaluating coverage and the set size of test data points. This raises the question of how to determine optimal $\mathbf{\omega}$ values in an intuitive and effective manner. To address this challenge, drawing inspiration from \cite{zargarbashi2023conformal, huang2023uncertainty}, we propose leveraging a model-based approach to learn the optimal parameters via a coverage and efficiency proxy loss. This methodology is further investigated in Section 4. 

%% file: 4algorithm.tex
\section{Methodology}
In this section, we present \method, a generic framework for addressing the challenges when the exchangeability condition is violated in temporal graphs. The key idea of the \method is integrating the theory shown in Lemma \ref{th: nonexchangeble CP} to address the non-exchangeability challenge. Additionally, we provide a topological and temporal non-conformity score to better capture uncertainty arising from changes in temporal graphs. Whereas, the bound provided in Lemma \ref{th: nonexchangeble CP} can be loose in certain cases, which compromises the coverage and efficiency. Thus, we offer an end-to-end framework to minimize the coverage gap while maintaining high efficiency.
Particularly, we first introduce the overall framework, followed by a detailed discussion on (M1) a topological and temporal non-conformity score, (M2) an efficiency-aware optimization algorithm. At last, we present an end-to-end optimization process to train \method effectively. 
\begin{figure*}[ht]
    \centering
    \includegraphics[width=\textwidth]{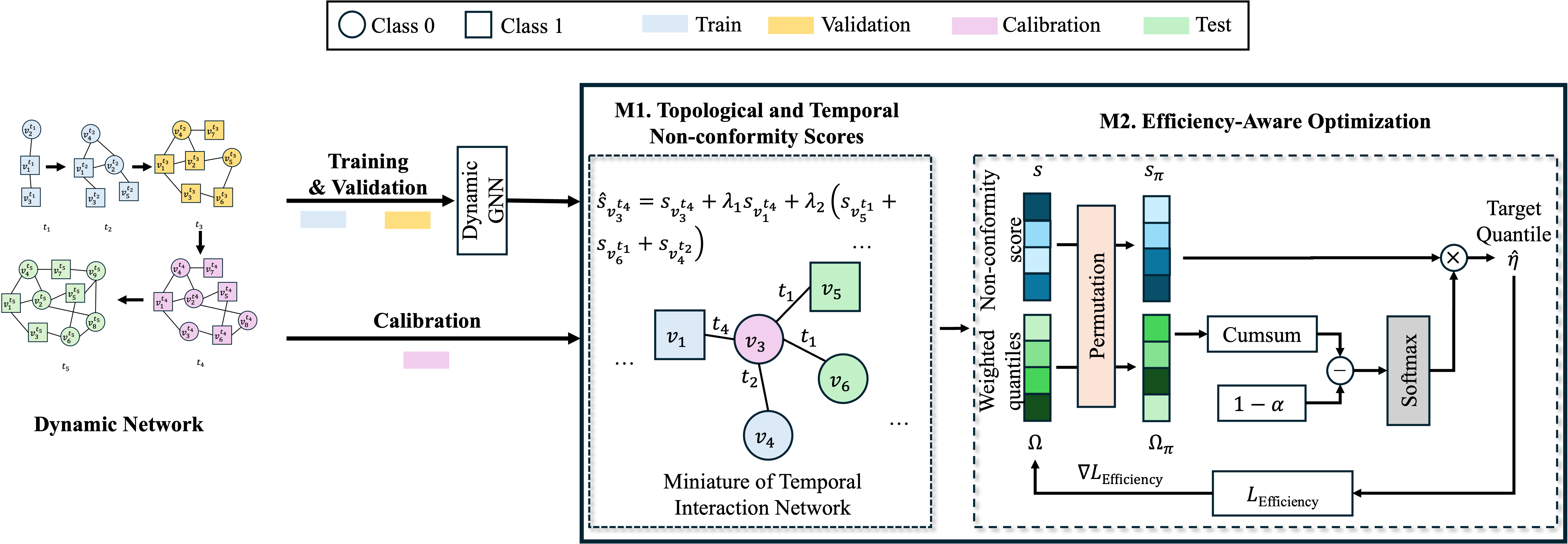}
    \captionsetup{skip=2pt}
    \caption{The Overview of our proposed framework \method, which is composed of two modules: (M1) topological and temporal diffusion-based non-conformity scores and (M2) efficiency-aware optimization.}
    \label{fig: overall framework}
\end{figure*}

\subsection{Framework Overview} \label{sec: framework overview}
Building upon the theoretical insights from Lemma \ref{th: nonexchangeble CP}, we introduce a novel algorithm named \method. This algorithm is tailored for temporal GNNs to leverage the non-exchangeability. In the original non-exchangeability theory \cite{barber2023conformal}, the authors choose TPS~\cite{Sadinle_2018} as the non-conformity score, and the weights utilized for quantile calculation are predetermined, lacking optimization for improved efficiency. Additionally, varying datasets may necessitate different weight settings, adding complexity when applied to diverse datasets. Our method addresses these limitations by employing optimized quantile calculation through a combination of topological and temporal diffusion non-conformity scores and learnable weighted parameters that are backward efficient aware. The overarching framework is illustrated in Fig. \ref{fig: overall framework}.
\vspace{1mm}

\noindent \textbf{M1. Non-Conformity Score Computation via Topological and Temporal Diffusion. }
The high-level idea of conformal prediction is to use a calibration method to provide a prediction set, which can guarantee that the probability of the real label lying in this set is at least as high as the desired value. In order to realize this calibration, a score called a non-conformity score is proposed to measure how unusual a sample is compared to a training sample. While many types of non-conformity scores are proposed, they end up with different performances when comparing the performance of coverage and efficiency of the conformal prediction sets. Besides, according to Lemma \ref{th: nonexchangeble CP}, the upper bound of the coverage gap is influenced by the non-conformity score. Furthermore, we argue that in temporal graph neural networks, each node's representation is affected not only by its topological neighbors but also by its temporal ones, which means that the temporal neighbors remain an influence on the current node. We define the temporal neighbors in temporal graphs as follows.
\begin{definition}[Temporal Graph Neighbors]
    Given a temporal node $v^t$ at timestamp $t$, the neighbors of the particular occurrence $v^t$ are termed as
    \begin{equation}
        \mathcal{N}_{v^t}=\{ v_i^t|f(v_i^{t_{i}}, v^t) \leq d_{st}, |t - t_i| \leq t_{st}, v_i^t \in \Tilde{\mathcal{V}} \},
    \end{equation}
    where $f(\cdot)$ denotes the shortest path between two nodes, and $d_{st}$ and $t_{st}$ are user-defined topological range threshold and temporal range threshold, respectively, to calculate the neighbors across the whole temporal graph.
\end{definition}
Inspired by \cite{pmlr-v202-h-zargarbashi23a}, we argue that in temporal graphs, not only do topological neighbors affect the distribution of non-conformity scores, but also the temporal neighbors. The intuition comes from the work of \cite{bahri2021locally}, where the authors argue that the true distribution can still be approached even if we only estimate it by using hard labels from the true distribution. Specifically for temporal graphs, the true distributions come from topological neighbors as well as temporal neighbors. Hence, we propose a topological and temporal non-conformity score to better reflect distributional variations across topological and temporal dimensions:
\begin{equation} \label{eq:non-conformity score}
    \hat{s}_i^{t_i} = (1 - \lambda_1 - \lambda_2)s_i^{t_i} + \frac{\lambda_1}{|N_i|}\sum_{j \in N_i}s_j + \frac{\lambda_2}{|N_i^{t_i}|}\sum_{j \in N_i^{t_i}}s_j^{t_i},
\end{equation}
where $\lambda_1$ and $\lambda_2$ denote how the non-conformity score would be affected via neighborhood nodes and temporal nodes. $N_i$ is the neighborhood nodes at time $t$ while $N_i^{t_i}$ is the temporal neighborhood nodes that are within a certain time before $t_i$ associated with node index $i$. We can start the diffusion process with initial non-conformity score $s_i^t = |y_i^t - f(\mathbf{x}_i^t)|$. Note that at each timestamp, the neighborhood is different, and the representation of each node also changes with time. Here, we introduce two parameters $\lambda_1$ and $\lambda_2$ to decide how much influence a node can get from its topological neighbors and temporal neighbors. In this paper, we use the grid search method to select the best parameters. However, our idea is that the parameters are chosen based on the dataset and even should be decided according to the node. The intuition is that different nodes may behave differently, they can either get more affected by their topological neighbors or their temporal neighbors. Thus, it is better to learn the parameters instead of giving a fixed value at the beginning. However, it is still an open question to research, and we will leave this to future work.

\noindent \textbf{M2. Efficiency-Aware Optimization Algorithm.} In standard conformal prediction, any non-conformity score can be used to construct prediction sets. However, different non-conformity scores exhibit varying levels of efficiency, and these scores do not inherently account for inefficiencies during quantile calculation, meaning optimization for efficiency is not automatically incorporated. 
Moreover, consistent with the theory of non-exchangeability, the coverage gap between non-exchangeable and exchangeable conditions depends on both the choice of weighted parameters and the specific non-conformity scores used. 
To obtain a more accurate quantile, we introduce a soft selection mechanism for determining the desired quantile value defined as follows:
\begin{equation} \label{eq: differentiable weighted quantiles}
    \begin{split}
        \mathbf{\Gamma} &= |cumsum(\Omega) - (1-\alpha)| \\
        \mathbf{B} &= \{\beta_i|\beta_i=\frac{e^{-\omega_i / T}}{\sum_{j=1}^{n} e^{-\omega_j / T}}\} \\
        \hat{\eta} &= S_\pi \mathbf{B}, 
    \end{split}
\end{equation}
where $T$ is a hyper-parameter that controls the soft assignment of the prediction set, $S_\pi=\{s_{\pi(1)},\ldots,s_{\pi(n)}\}$ is the sorted non-conformity score, and $\mathbf{\Omega}=\{\omega_1,\ldots,\omega_n\}$ is the weighted parameters.

Based on a differential selection of the desired quantile value, to ensure both desired coverage and improved efficiency, we propose a method to optimize the weighted parameters based on coverage and efficiency considerations without necessitating changes to the model training process. To optimize the weighted parameters for efficiency, it's essential to have a suitable proxy that can simulate the size of the prediction set. Thus, given a calibration training dataset $D_{c\_train}$ and a calibration validation dataset $D_{c\_valid}$, we define the non-conformity score for each class $k$ as $s(f(\tilde{G}, \tilde{X}, \tilde{\mathcal{\theta}}, \mathbf{x}_i), y_k)$, where $s(\cdot)$ represents the topological and temporal-aware non-conformity score. Consequently, the efficiency loss is defined as:
\begin{equation} \label{eq:efficiency loss}
    \mathcal{L}_\text{Efficiency} = \sum_{i=1}^n \sum_{k=1}^K \sigma(\frac{s(f(\tilde{G}, \tilde{X}, \tilde{\mathcal{\theta}}, \mathbf{x}_i), k) - \hat{\eta}}{\tau}),
\end{equation}
where $K$ is the number of classes, $\sigma$ is the sigmoid function, $\hat{\eta}$ is the differentiable quantile we get from Eq. \ref{eq: differentiable weighted quantiles}, and $\tau$ is a hyper-parameter that controls the assignment results.

\subsection{Optimization}
\begin{algorithm}
    \caption{Training with efficiency loss}
    \label{alg:model training}
    \begin{algorithmic}[1]
        \State{\textbf{Input: } (i) An input temporal graph model $f(\tilde{\mathcal{G}}, \tilde{\mathbf{X}}, \tilde{\theta})$ with a calibration training dataset $D_{c\_train}$ and corresponding labels $\tilde{\mathbf{y}}$. (ii) initialized parameters $\Omega=\{\omega_1,...,\omega_k \}$}
        
        \State{\textbf{Output: } Optimized weighted quantiles parameters $\hat{\Omega}=\{\hat{\omega}_1,...,\hat{\omega}_k \}$}
        
        \For {$epoch = 1 \rightarrow N$}
            \For {$v_i \in D_{c\_train}$ }
                \State Computer model output $\hat{y_i}=f(\tilde{\mathcal{G}}, \tilde{x}_i, \tilde{\theta})$
                \State Compute the non-conformity score based on Eq. \ref{eq:non-conformity score}.
                \State Compute weighted quantiles through Eq. \ref{eq: differentiable weighted quantiles}.
                \State Calculate the efficicency loss through Eq. \ref{eq:efficiency loss}.
                \State Optimized loss using backward propagation.
            \EndFor
        \EndFor
    \end{algorithmic}
\end{algorithm}
To enhance both efficiency and achieve the desired coverage in temporal GNNs, we introduce a method to optimize the weighted parameters within the Lemma \ref{th: nonexchangeble CP}. The overall framework comprises several key steps and we provide the pseudo-code of the training process in Algorithm \ref{alg:model training}. Initially, we start with a standard training process to develop a temporal GNN using the training dataset. This model is denoted as $f$ and can be any temporal GNN. Then, we compute the topological and temporal non-conformity score using Eq. \ref{eq:non-conformity score}. Recall that this non-conformity score is calculated as long as there is a valid temporal graph model, which means that the non-conformity score is fixed after the training of the temporal GNN. The efficiency loss can be calculated based on Eq.~\ref{eq:efficiency loss} given the differential quantile $\hat{\eta}$ and the non-conformity score from Eq. \ref{eq:non-conformity score} on the calibration dataset(Section ~\ref{sec: experimental details}). Finally, the parameters in non-exchangeable conformal prediction are optimized in an end-to-end process using the calibration training dataset.

%% file: 5experiments.tex
\section{Experiments}

\begin{table*}[h]
    \small
    \centering
    \setlength{\tabcolsep}{0pt}
    \captionsetup{skip=2pt}
    \caption{Experimental results over four datasets. The number in green bold indicates the best performance when coverage is satisfied, while the blue underline indicates the second-best performance. \greencheck~ indicates that the calibration method reaches the target coverage (95\%) while \redmark~ indicates the opposite. \method w/o $\omega$ denotes \method without weighted quantiles optimization, \method w/o s denotes \method without topological and temporal non-conformity score. OOM here indicates that the method is out of memory in our machine and we are not able to get the necessary experimental results we need.}
    \begin{tabular}{c|cc|cc|cc|cc|cc|cc}
    \hline
        \multirow{3}{*}{Methods} & \multicolumn{6}{c|}{WIKI} & \multicolumn{6}{c}{REDDIT} \\
        \cline{2-13}
        ~ & \multicolumn{2}{c|}{TGAT} & \multicolumn{2}{c|}{JODIE} & \multicolumn{2}{c|}{TGN} & \multicolumn{2}{c|}{TGAT} & \multicolumn{2}{c|}{JODIE} & \multicolumn{2}{c}{TGN} \\
        \cline{2-13}
        ~ & Coverage $\uparrow$ & Efficiency $\downarrow$ & Coverage $\uparrow$ & Efficiency $\downarrow$ & Coverage $\uparrow$ & Efficiency $\downarrow$ & Coverage $\uparrow$ & Efficiency $\downarrow$ & Coverage $\uparrow$ & Efficiency $\downarrow$ & Coverage $\uparrow$ & Efficiency $\downarrow$ \\ \hline
        TPS & 0.89±0.01\redmark & 1.89±0.01 & 0.68±0.10\redmark & 1.68±0.11 & 0.56±0.06\redmark & 1.56±0.06 &0.61±0.15\redmark & 1.61±0.15 & 0.63±0.05\redmark & 1.63±0.05 & 0.67±0.10\redmark & 1.67±0.11 \\
        APS & 0.88±0.09\redmark & 1.87±0.10 & 0.84±0.11\redmark & 1.82±0.12 & 0.90±0.17\redmark & 1.88±0.20 & 0.91±0.09\redmark & 1.91±0.09 & 0.85±0.07\redmark & 1.83±0.08 & 0.90±0.08\redmark & 1.89±0.08 \\ 
        RAPS & 0.99±0.01\greencheck & 1.76±0.15 & 1.00±0.00\greencheck & 1.70±0.14 & 1.00±0.00\greencheck & 1.81±0.14 & 1.00±0.00\greencheck & 1.65±0.15 & 1.00±0.00\greencheck & 1.63±0.04 & 0.99±0.01\greencheck & 1.70±0.16 \\ 
        DAPS & 0.86±0.02\redmark & 1.36±0.11 & 0.98±0.00\greencheck & 1.53±0.07 & 0.85±0.03\redmark & 1.45±0.04 & 0.88±0.04\redmark & 1.66±0.07 & 0.95±0.01\greencheck & 1.73±0.06 & 0.88±0.06\redmark & 1.84±0.07 \\ \hline
        CF-GNN & 1.00±0.00\greencheck & 1.72±0.23 & 1.00±0.00\greencheck & 1.69±0.11 & 1.00±0.00\greencheck & 1.57±0.28 & 1.00±0.00\greencheck & \underline{\textcolor{blue}{1.21±0.07}} & 0.91±0.02\redmark & 1.84±0.13 & 0.94±0.03\redmark & 1.71±0.12 \\ \hline
        NEX & 1.00±0.00\greencheck & 1.99±0.01 & 1.00±0.00\greencheck & 1.81±0.03 & 1.00±0.00\greencheck & 1.76±0.13 & 1.00±0.00\greencheck & 1.76±0.11 & 1.00±0.00\greencheck & 1.65±0.05 & 1.00±0.00\greencheck & 1.73±0.10 \\ 
        NAPS & 1.00±0.00\greencheck & 1.86±0.00 & 0.95±0.01\greencheck & 1.56±0.03 & 0.95±0.01\greencheck & 1.83±0.01 & 1.00±0.00\greencheck & 2.00±0.00 & 1.00±0.00\greencheck & 1.75±0.25 & 1.00±0.00\greencheck & 2.00±0.00 \\ \hline
        UGNN & 1.00±0.00\greencheck & 1.97±0.00 & 0.96±0.01\greencheck & 1.30±0.48 & 1.00±0.00\greencheck & 1.97±0.00 & 0.99±0.00\greencheck & 1.98±0.00 & 0.97±0.01\greencheck & \underline{\textcolor{blue}{1.43±0.48}} & 0.99±0.00\greencheck & 1.98±0.00\\ \hline
        \method w/o $\omega$ & 1.00±0.00\greencheck & 1.57±0.05 & 1.00±0.00\greencheck & 1.47±0.26 & 1.00±0.00\greencheck &	1.61±0.33 & 1.00±0.00\greencheck & 1.47±0.11 & 1.00±0.00\greencheck & 1.55±0.06 & 1.00±0.00\greencheck &	1.55±0.16 \\
        \method w/o s & 1.00±0.00\greencheck & \underline{\textcolor{blue}{1.49±0.08}} & 1.00±0.00\greencheck & \underline{\textcolor{blue}{1.25±0.09}} & 1.00±0.00\greencheck & \underline{\textcolor{blue}{1.27±0.05}} & 1.00±0.00\greencheck & 1.53±0.12 & 0.99±0.01\greencheck & 1.62±0.07 & 0.97±0.02\greencheck & \textbf{\textcolor{darkpastelgreen}{1.23±0.18}} \\
        \method & 0.97±0.03\greencheck & \textbf{\textcolor{darkpastelgreen}{1.31±0.14}} & 0.98±0.03\greencheck & \textbf{\textcolor{darkpastelgreen}{1.16±0.14}} & 0.99±0.01\greencheck & \textbf{\textcolor{darkpastelgreen}{1.17±0.12}} & 0.97±0.02\greencheck & \textbf{\textcolor{darkpastelgreen}{1.07±0.12}} & 0.97±0.02\greencheck & \textbf{\textcolor{darkpastelgreen}{1.16±0.15}} & 0.95±0.00\greencheck & \underline{\textcolor{blue}{1.29±0.10}} \\ \hline
        \hline

        \multirow{3}{*}{Methods} & \multicolumn{6}{c|}{DBLP} & \multicolumn{6}{c}{IBM} \\
        \cline{2-13}
        ~ & \multicolumn{2}{c|}{TGAT} & \multicolumn{2}{c|}{JODIE} & \multicolumn{2}{c|}{TGN} & \multicolumn{2}{c|}{TGAT} & \multicolumn{2}{c|}{JODIE} & \multicolumn{2}{c}{TGN} \\
        \cline{2-13}
        ~ & Coverage$\uparrow$ & Efficiency$\downarrow$ & Coverage$\uparrow$ & Efficiency$\downarrow$ & Coverage$\uparrow$ & Efficiency$\downarrow$ & Coverage$\uparrow$ & Efficiency$\downarrow$ & Coverage$\uparrow$ & Efficiency$\downarrow$ & Coverage$\uparrow$ & Efficiency$\downarrow$ \\ \hline

        TPS & 1.00±0.00\greencheck & 3.17±0.07 & 1.00±0.00\greencheck & 3.50±0.11 & 1.00±0.00\greencheck & 3.59±0.33 & 0.76±0.03\redmark & 1.76±0.03 & 0.72±0.06\redmark & 1.71±0.06 & 0.77±0.06\redmark & 1.77±0.05  \\ 
        APS & 1.00±0.00\greencheck & 3.52±0.23 & 1.00±0.00\greencheck & 3.43±0.07 & 1.00±0.00\greencheck & 3.49±0.18 & 0.90±0.03\redmark & 1.89±0.03 & 0.86±0.03\redmark & 1.85±0.03 & 0.86±0.03\redmark & 1.85±0.03 \\ 
        RAPS & 0.96±0.01\greencheck & 3.48±0.14 & 0.97±0.01\greencheck & 3.45±0.09 & 0.96±0.01\greencheck & 3.47±0.14 & 1.00±0.00\greencheck & 1.81±0.05 & 0.99±0.01\greencheck & 1.73±0.06 & 1.00±0.00\greencheck & 1.77±0.06 \\
        DAPS & 0.93±0.02\redmark & 3.22±0.22 & 0.95±0.01\greencheck & 3.42±0.33 & 0.93±0.01\redmark & 3.23±0.31 & 0.98±0.02\greencheck & 1.95±0.05 & 0.92±0.00\redmark & 1.82±0.00 & 0.95±0.01\greencheck & 1.82±0.01 \\ \hline
        CF-GNN & 0.99±0.00\greencheck & 4.64±0.19 & 0.99±0.01\greencheck & 4.59±0.33 & 0.99±0.01\greencheck & 4.59±0.25 & 1.00±0.00\greencheck & 1.11±0.31 & 1.00±0.00\greencheck & 1.46±0.38 & 1.00±0.00\greencheck & 1.40±0.51 \\ \hline
        NEX & 0.97±0.01\greencheck & 3.66±0.19 & 0.97±0.01\greencheck & 3.68±0.21 & 0.97±0.01\greencheck & 3.75±0.30 & 1.00±0.00\greencheck & 1.66±0.07 & 0.99±0.01\greencheck & 1.61±0.11 & 1.00±0.00\greencheck & 1.67±0.07 \\ 
        NAPS & 1.00±0.00\greencheck & 4.68±0.04 & 1.00±0.00\greencheck & 4.24±0.04 & 1.00±0.00\greencheck & 4.79±0.04 & 1.00±0.00\greencheck & 1.58±0.04 & 1.00±0.00\greencheck & 1.45±0.04 & 0.99±0.01\greencheck & 1.22±0.03\\ \hline
        UGNN & 0.90±0.01\redmark & 2.90±0.52 & 0.94±0.04\redmark & 4.15±0.67 & 0.98±0.02\greencheck & 4.88±0.11 & OOM & OOM & OOM & OOM & OOM & OOM \\ \hline
        \method w/o $\omega$ & 0.96±0.01\greencheck & 3.44±0.41 & 0.96±0.01\greencheck & 3.19±0.06 & 0.95±0.00\greencheck & \textbf{\textcolor{darkpastelgreen}{3.07±0.11}} & 0.96±0.01\greencheck & 1.24±0.11 & 0.97±0.01\greencheck & 1.57±0.16 & 0.96±0.01\greencheck & 1.43±0.08 \\
        \method w/o s & 0.95±0.00\greencheck & \textbf{\textcolor{darkpastelgreen}{3.12±0.13}} & 0.96±0.00\greencheck & \underline{\textcolor{blue}{3.13±0.07}} & 0.95±0.01\greencheck & 3.33±0.15 & 0.95±0.00\greencheck & \underline{\textcolor{blue}{1.08±0.10}} & 0.95±0.00\greencheck & \underline{\textcolor{blue}{1.22±0.15}} & 0.95±0.00\greencheck & \underline{\textcolor{blue}{1.08±0.09}} \\
        \method & 0.95±0.01\greencheck & \underline{\textcolor{blue}{3.14±0.23}} & 0.96±0.00\greencheck & \textbf{\textcolor{darkpastelgreen}{2.98±0.09}} & 0.96±0.01\greencheck & \underline{\textcolor{blue}{3.24±0.23}} & 0.98±0.00\greencheck & \textbf{\textcolor{darkpastelgreen}{1.02±0.01}} & 0.99±0.01\greencheck & \textbf{\textcolor{darkpastelgreen}{1.01±0.01}} & 0.99±0.00\greencheck & \textbf{\textcolor{darkpastelgreen}{1.01±0.01}} \\ \hline
    \end{tabular}
    \label{tab:experimental results}
    \vspace{-1em}
\end{table*}

In this section, we analyze the following key aspects to demonstrate the effectiveness of \method: (i) we evaluate the performance of \method on four benchmark temporal graph datasets where \method exhibits superior performance compared to other baselines (Section \ref{sec: effectiveness}). (ii) we conduct ablation studies to demonstrate the necessity of each module in \method and show how mis-coverage level and training data size affect \method's performance (Section \ref{sec: ablation study}); (iii) we report the parameter analysis on the topological and temporal non-conformity score in Section~\ref{sec: parameter analysis} and scalability test in Section~\ref{sec: scalability test} to show that \method is scalable and can achieve convincing performance with minimum tuning efforts; (iv) we offer a case study demonstrating how \method enhances the performance of a standard temporal GNN by generating prediction sets that more closely adhere to the desired coverage level (Section.\ref{sec: case study}).

\subsection{Experimental Settings} \label{sec: experimental details}
\noindent\textbf{Experimental Setup.} We adhere to a standard procedure for training the node classification model. Each dataset is divided into train/validation/calibration train/calibration validation/test datasets in the proportions of 50\%, 10\%, 10\%, 10\%, and 20\%. To mitigate the influence of parameter optimization randomness, we conduct multiple runs for each backbone. The implementation of JODIE, TGAT, and TGN follows the work from \cite{zhou2022tgl}. The implementation of the non-conformity score follows the work from the original paper TPS \cite{Sadinle_2018}, APS \cite{romano2020classification}, and RAPS \cite{angelopoulos2020uncertainty}. All codes in this paper are programmed in Python 3.10.13 and PyTorch 2.2.1. All experiments are performed on a Linux server with 64 AMD EPYC 7313 CPUs and 1 Nvidia Tesla A100 SXM4 GPU with 80 GB memory.

\noindent \textbf{Baselines.} We compare \method against several baseline models spanning four key categories: (1) non-conformity score based approaches, including TPS~\cite{Sadinle_2018}, APS~\cite{romano2020classification}, and RAPS~\cite{angelopoulos2020uncertainty}, DAPS~\cite{zargarbashi2023conformal}; (2) GNN based approaches, including CF-GNN~\cite{huang2023uncertainty}; (3) non-exchangeable based approaches including NEX~\cite{barber2023conformal} and NAPS~\cite{clarkson2023distribution}; (4) stochastic block model based approaches, including unfolding GNN (UGNN)~\cite{davis2024valid}. The details for each baseline model are introduced in Appendix \ref{ap: baselines introduction}. Finally, to assess the generalizability of \method, we evaluate its performance across three widely adopted temporal GNN models: JODIE~\cite{kumar2019predicting}, TGAT~\cite{xu2020inductive}, and TGN~\cite{rossi2020temporal}. While some baselines, such as CF-GNN and NAPS, are designed for static GNNs, prior work~\cite{radstok2021leveraging} shows that temporal graphs can be converted into static ones for calibration or representation learning. Thus, for the NAPS baseline, we use the transformed static graphs for evaluation. We adapt CF-GNN by extracting temporal node embeddings with a backbone temporal GNN, then building a static graph where each node-time pair is a unique node connected by observed temporal interactions. We also evaluated CF-GNN without the graph structure and achieved the best fairness results.

\noindent \textbf{Datasets.} To evaluate the effectiveness of \method, we conduct experiments on four diverse real-world datasets: WIKI~\cite{kumar2019predicting}, REDDIT~\cite{kumar2019predicting}, DBLP~\cite{hu2018developing}, and the IBM Anti-Money Laundering dataset~\cite{altman2024realistic}. These datasets represent a diverse range of real-world application scenarios, ensuring a comprehensive assessment of our proposed approach. A detailed description of each dataset is available in Appendix~\ref{ap: data statistics}, with statistical summaries in Table~\ref{tab: statistics of datasets} and temporal characteristics in Table~\ref{tab: temporal statistics}.

\noindent\textbf{Evaluation Metrics.} To rigorously evaluate the performance of \method, we employ two fundamental evaluation metrics: \textit{coverage} and \textit{efficiency}. The \textit{coverage} metric quantifies the reliability of the uncertainty estimates by measuring the proportion of instances where the ground truth label is included within the predicted set. In contrast, the \textit{efficiency} metric assesses the conciseness of the prediction set, indicating the models' ability to generate informative and precise predictions. The definitions are as follows:  
\begin{equation} \label{eq:coverage}
    \text{coverage} := \frac{1}{|\mathcal{D}_{t}|} \sum_{i \in \mathcal{D}_{t}} \mathbbm{1}(y_i \in C_i),
\end{equation}
\begin{equation} \label{eq:efficiency}
    \text{efficiency} := \frac{1}{|\mathcal{D}_{t}|} \sum_{i \in \mathcal{D}_{t}} |C_i|,
\end{equation}
where $C_i$ is the prediction set for a given data point, and $y_i$ is the corresponding ground truth label. There is an inherent trade-off between coverage and efficiency. Higher coverage can be achieved by increasing the quantile value, but this enlarges the prediction set, reducing specificity and discriminative power. Finding the right balance between these metrics is essential for reliable and practical uncertainty quantification.

\subsection{Comparison Experimental Results.} \label{sec: effectiveness}
Our experiments, as shown in Table~\ref{tab:experimental results}, reveal that \method consistently achieves the pre-defined mis-coverage across all datasets and backbone models and further demonstrates superior efficiency in generating conformal prediction sets for the majority of datasets and backbone models. We also observe that all the non-conformity scores fail to achieve the pre-defined coverage level except for RAPS. A simple guess for this phenomenon is that RAPS contains regularization parameters based on the probability distribution, which mitigate the influence of distribution shift and unreliable small distributions. Moreover, we find that methods based on the exchangeability assumption, such as all the non-conformity scores, CF-GNN, and UGNN, fail to achieve pre-defined coverage in some datasets over some temporal GNNs. For instance, the UGNN method achieves the 95\% percent coverage level at the WIKI and REDDIT datasets and fails in the DBLP dataset. Additionally, the UGNN method requires so much memory that it fails to output valid results for larger temporal datasets like IBM. However, methods that required no assumption on exchangeability, such as NEX, NAPS, and \method, all achieve the pre-defined coverage across all the datasets and backbone temporal GNNs. This suggests that in temporal graphs, the exchangeability condition does not hold in every situation, and the methods that cover both exchangeable and non-exchangeable conditions perform better. Compared between NEX, NAPS, and \method, we reach the conclusion that \method outperforms in the efficiency metric. For example, in the WIKI dataset, when evaluating the \method with three different backbone models, we observe efficiency improvements of 34.2\%, 35.9\%, and 33.5\% for the TGAT, JODIE, and TGN models compared to NEX, respectively. This observation proves the effectiveness of our proposed framework in achieving higher efficiency.

\subsection{Ablation Study} \label{sec: ablation study}
\textbf{Effectiveness of \method Modules.} To rigorously evaluate the effectiveness of our framework's components, we conduct comprehensive ablation experiments across multiple runs shown in Table~\ref{tab:experimental results}. \method denotes the full functionality, \method w/o s denotes \method without the topological and temporal non-conformity score (M1), and \method w/o $\omega$ denotes \method without weighted quantiles optimization (M2).  Our findings indicate that the efficiency performance follows the order: \method > \method w/o s > \method w/o $\omega$ in general. For example, in the WIKI dataset, \method w/o $\omega$ shows efficiency improvements of 21.1\%, 18.8\%, and 8.5\% for the TGAT, JODIE, and TGN models, respectively. In contrast, \method w/o s achieves efficiency increases of 25.1\%, 30.9\%, and 27.8\% for TGAT, JODIE, and TGN compared to the non-exchangeable baseline. These ablation results prove the necessity of our modules.
\begin{figure*}[h!]
    \centering
    \begin{subfigure}[b]{0.23\textwidth}
        \centering
        \includegraphics[width=\textwidth]{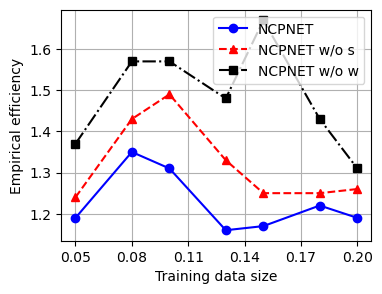}
        \captionsetup{skip=2pt}
        \caption{Efficiency vs Data Size}
        \label{fig:training data size influence on efficicency}
    \end{subfigure}
    \hfill
    \begin{subfigure}[b]{0.23\textwidth}
        \centering
        \includegraphics[width=\textwidth]{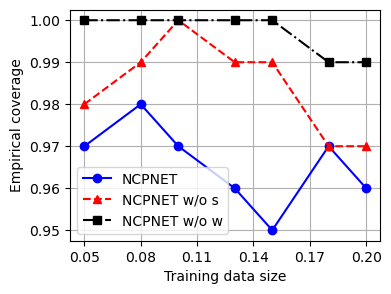}
        \captionsetup{skip=2pt}
        \caption{Coverage vs Data Size}
        \label{fig:training data size influence on coverage}
    \end{subfigure}
    \hfill
    \begin{subfigure}[b]{0.23\textwidth}
        \centering
        \includegraphics[width=\textwidth]{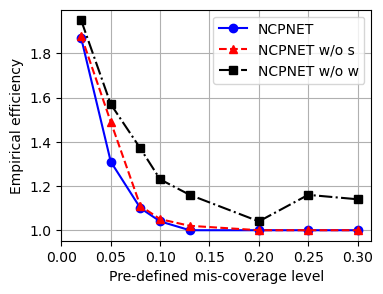}
        \captionsetup{skip=2pt}
        \caption{Efficiency vs Pre-defined $\alpha$}
        \label{fig:parameter analysis on alpha}
    \end{subfigure}
    \hfill
    \begin{subfigure}[b]{0.23\textwidth}
        \centering
        \includegraphics[width=\textwidth]{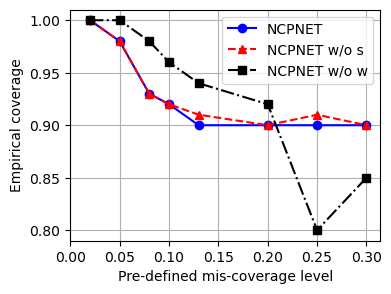}
        \captionsetup{skip=2pt}
        \caption{Coverage vs Pre-defined $\alpha$}
        \label{fig:parameter analysis coverage on alpha}
    \end{subfigure}
    \captionsetup{skip=2pt}
    \caption{Efficiency and coverage on various training data sizes and mis-coverage level. Fig. \ref{fig:training data size influence on efficicency} and Fig. \ref{fig:training data size influence on coverage}: \method's efficiency and coverage performance on different training data size, Fig. \ref{fig:parameter analysis on alpha} and Fig. \ref{fig:parameter analysis coverage on alpha}: \method's efficiency and coverage performance on different mis-coverage levels.}
    \label{fig:parameter analysis}
\end{figure*}

\noindent \textbf{Training Data Size and Mis-coverage Level}. To further test the influence of various parameters in \method. We select parameters such as pre-defined mis-coverage $\alpha$, and training data size to test the performance change based on various parameter ranges. Overall, our experiments indicate that the efficiency and coverage tend to stay the same when calibration training data size increases, as shown in Fig. \ref{fig:training data size influence on efficicency} and Fig. \ref{fig:training data size influence on coverage}, which suggests that a relatively small calibration set is sufficient to output the best performance in \method. Additionally, Fig.~\ref{fig:training data size influence on efficicency} shows that NCPNET maintains stable efficiency even with small calibration sets. Efficiency improves slightly with more data, but gains plateau due to regularization from learned M2 weights. In Fig.~\ref{fig:parameter analysis on alpha} and Fig.~\ref{fig:parameter analysis coverage on alpha}, even though the mis-coverage level increases, \method still achieves the coverage levels over all ranges.
\begin{figure}[h!]
    \centering
    \includegraphics[width=0.45\textwidth]{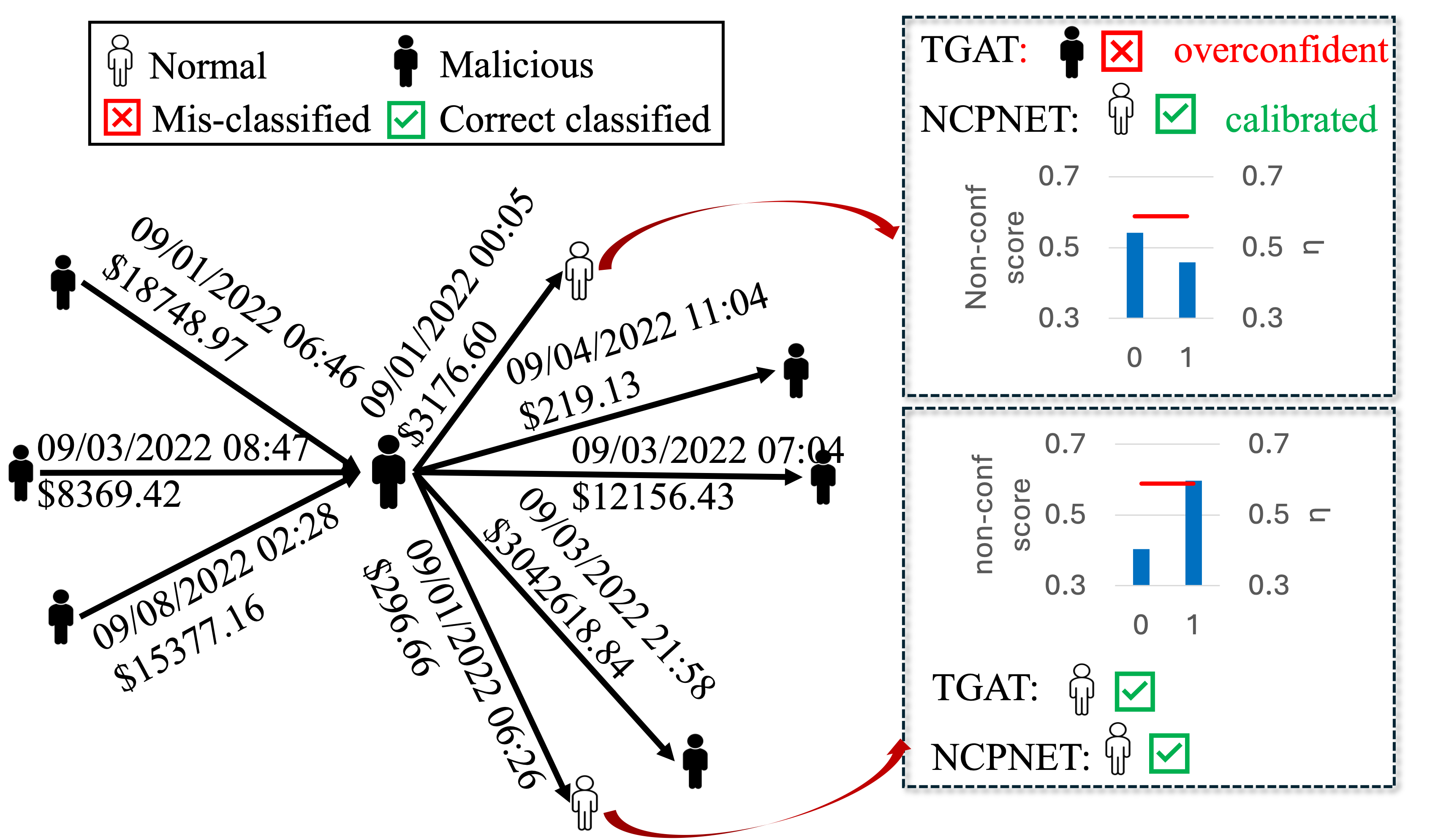}
    \captionsetup{skip=2pt}
    \caption{Case study on IBM anti-money laundering dataset \cite{altman2024realistic}. The red cross indicates the misclassified result, while the green check indicates the correct result.}
    \label{fig:case study}
\end{figure}

\subsection{Scalability Analysis} \label{sec: scalability test}
\begin{figure}[h]
    \begin{subfigure}{0.45\linewidth}
        \centering
        \includegraphics[width=\textwidth]{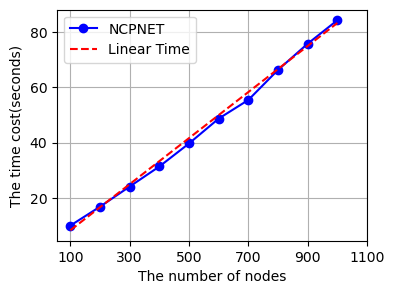}
        \caption{Test on the node number}
        \label{fig: scalability test on the node number}
    \end{subfigure}
    \qquad
    \begin{subfigure}{0.45\linewidth}
        \centering
        \includegraphics[width=\textwidth]{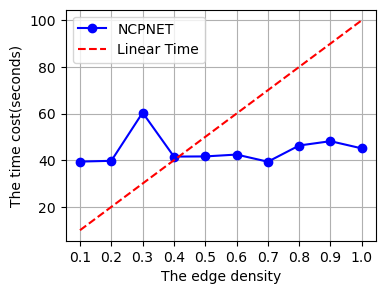}
        \caption{Test on the edge density}
        \label{fig: scalability test on the edge density}
    \end{subfigure}
    \captionsetup{skip=2pt}
    \caption{Scalability test on the node number and edge density}
\end{figure}

In this analysis, we evaluate the scalability of \method by measuring its computational efficiency across varying training graph node numbers and edge density. We measure the training time over multiple runs while systematically increasing both the number of nodes and the edge density of synthetic data to assess the model's practical applicability in real-world scenarios. The empirical results, shown in Fig.~\ref{fig: scalability test on the node number}, demonstrate that the computational overhead grows near-linearly with the number of nodes, indicating efficient scaling of our method.
The empirical results, shown in Fig.~\ref{fig: scalability test on the edge density}, demonstrate that the computational overhead nearly stays the same when the edge density increases, given a fixed node number. These patterns from the results suggest that our model can effectively handle the increasing complexity of more complicated graphs without a prohibitive computational burden.

\subsection{Parameter Sensitivity Analysis} \label{sec: parameter analysis}
We also examine the influence of different diffusion parameters in Eq.~\ref{eq:non-conformity score}. We select the DBLP dataset and TGAT as the backbone temporal GNN model to conduct the parameter analysis test. We chose this dataset because the number of nodes changes rapidly at different time steps. Thus, it is a representative dataset to show the influence of the parameters. Test results are shown in Table~\ref{tab: parameter analysis on diffusion} on diffusion parameters; we can see that increasing $\lambda_2$, which controls the temporal neighbor's contribution to the non-conformity score, tends to decrease the prediction set with the sacrifice of coverage. In terms of $\lambda_1$, which controls the contribution of structural neighbors, the parameter's increase also tends to decrease the prediction set size at the sacrifice of coverage. The choice of these two parameters is to set both of these parameters on a small scale($\lambda_1=\lambda_2=0.01$), which adds a little influence from structural and temporal neighbors. Due to the computational cost, we perform a limited grid search, but we find that the parameter settings consistently perform well across datasets and models.
\begin{table}[h!]
    \centering
    \setlength{\tabcolsep}{1pt}
    \captionsetup{skip=2pt}
    \caption{Parameter Analysis for Topological and Temporal Diffusion Non-conformity Score}
    \begin{tabular}{cccc|cccc}
        \hline
        $\lambda_1$ & $\lambda_2$ & Coverage & Efficiency & $\lambda_1$ & $\lambda_2$ & Coverage & Efficiency\\ \hline
        0.00 & 0.00 & 0.9384 & 3.4586 & 0.05 & 0.01 & 0.9318 & 3.3186 \\
        0.01 & 0.01 & 0.9503 & 3.4524 & 0.01 & 0.05 & 0.9318 & 3.3054 \\
        0.02 & 0.01 & 0.9484 & 2.9498 & 0.1 & 0.1 & 0.9381 & 3.2268 \\
        0.03 & 0.01 & 0.9456 & 2.9393 & 0.1 & 0.5 & 0.9323 & 2.8837 \\
        0.04 & 0.01 & 0.9463 & 2.9442 & 0.5 & 0.5 & 0.9337 & 2.8546 \\
        0.03 & 0.02 & 0.9394 & 3.4250 & 0.5 & 0.1 & 0.9492 & 3.1107 \\
        0.02 & 0.02 & 0.9326 & 3.4920 & 0.00 & 1.00 & 0.9332 & 3.4021 \\
        0.04 & 0.02 & 0.9414 & 3.0335 & 1.00 & 0.00 & 0.9451 & 3.0279 \\ \hline
    \end{tabular}
    \label{tab: parameter analysis on diffusion}
\end{table}
\vspace{-4mm}
\subsection{Case Study on Money Laundering Detection} \label{sec: case study}
In real-world applications, the confidence of deep learning models in their predictions is not always assured, especially when encountering out-of-distribution data that may compromise model performance. To more effectively evaluate model efficacy, we conduct a case study utilizing conformal prediction. In this study, we demonstrate that low-confidence outputs from the deep learning model are frequently linked to inaccuracies. We illustrate this using the IBM transactions dataset focused on anti-money laundering. As depicted in Fig. \ref{fig:case study}, normal accounts involved in transactions with malicious accounts may be more likely to be incorrectly flagged as malicious, potentially leading to the misclassification of innocent accounts. By implementing our \method approach, we can generate conformal sets that accurately represent the model's confidence in its predictions. This capability allows us to reduce misclassification rates and provide more reliable outputs. These advancements are particularly crucial in real-world contexts, especially in domains such as the detection of malicious transactions in finance, where overly confident predictions can lead to erroneous risk assessments.

%% file: 6relatedworks.tex
\section{Related Work}
\textbf{Conformal Prediction.} Vovk \cite{vovk2005algorithmic, vovk2017nonparametric} first introduced conformal prediction, which provides a guaranteed confidence level for prediction sets based on a predefined coverage rate. Since then, there have been many studies to improve the application~\cite{lu2022fair, fannjiang2022conformal} and development of the theory of conformal prediction~\cite{tibshirani2019conformal, xu2021conformal,wu2025bridging}. Given its robustness, enhancing efficiency is a primary objective when applying conformal prediction to various domains \cite{straitouri2023improving, ndiaye2022stable}. An effective non-conformity score can significantly improve the efficiency of conformal prediction sets under the exchangeability guarantee. Notable methods like TPS \cite{Sadinle_2018}, APS \cite{romano2020classification}, and RAPS \cite{angelopoulos2020uncertainty} enhance the efficiency of conformal prediction sets by employing different approaches to calculate the non-conformity score. 
Recently, \cite{zargarbashi2023conformal} proposed a diffusion-based non-conformity score that considers the topological structure when applying conformal prediction to graphs. Additionally, \cite{huang2023uncertainty} introduced a calibration GNN model aimed at achieving better-calibrated GNNs, ensuring improved efficiency and adherence to predefined coverage guarantees. Our work differs from these as we concentrate on the temporal graphs, where the exchangeability assumption is violated. We propose a topological and temporal diffusion-based non-conformity and an efficiency-aware optimization method to achieve better efficiency and empirical coverage guarantees.

\noindent \textbf{Temporal Graph Neural Networks.} 
Temporal graphs are often modeled as interaction streams, making it critical to capture latent evolution patterns in various domains. Existing methods~\cite{pareja2020evolvegcn, wang2024evolunet} can be classified into memory-based, GNN-based, RNN-based, and hybrid methods combining GNNs and RNNs. JODIE~\cite{kumar2019predicting} is a typical representative of memory-based methods, while TGAT~\cite{xu2020inductive} represents GNN-based methods. Additionally, methods like TGN~\cite{rossi2020temporal} combine memory blocks with GNN structures to capture both temporal and topological information. Other approaches, such as DySAT \cite{sankar2020dysat}, use GNNs to extract spatial features and then employ RNNs to capture temporal interactions based on the GNN-derived spatial representations. Recently, \cite{cong2023we} introduces the GraphMixer model that utilizes an MLP-mixer \cite{tolstikhin2021mlp, yu2022s2} to summarize temporal link information. We primarily focus on enabling conformal prediction for temporal GNN models. Our method is model-agnostic and can be applied to various types of temporal GNNs.

%% file: 7conclusion.tex
\section{Conclusion}


In this paper, we introduce an algorithm called \method, specifically developed to apply conformal prediction to temporal graph neural networks while accounting for both topological structures and temporal interactions. Our main objective is to incorporate non-exchangeability theory into the framework of temporal graphs, acknowledging that the exchangeability assumption is frequently violated due to time dependencies. We perform a theoretical analysis of temporal graphs and propose an upper bound to effectively address the gap in achieving the desired predefined coverage. We argue that a robust non-conformity score, which incorporates topological and temporal interactions among nodes, enhances efficiency, and we further advocate for efficiency-aware optimization to produce more effective conformal prediction sets. Extensive experiments on real-world datasets demonstrate that our algorithm significantly surpasses baseline methods in both efficiency and adherence to desired coverage levels.

%% file: ack.tex
\section*{Acknowledgements}
We thank the anonymous reviewers for their constructive comments. This work is supported by the National Science Foundation under Award No. IIS-2339989 and No. 2406439, DARPA under contract No. HR00112490370 and No. HR001124S0013, U.S. Department of Homeland Security under Grant Award No. 17STCIN00001-08-00,  Amazon-Virginia Tech Initiative for Efficient and Robust Machine Learning, Amazon AWS, Google, Cisco, 4-VA, Commonwealth Cyber Initiative, National Surface Transportation Safety Center for Excellence, and Virginia Tech. The views and conclusions are those of the authors and should not be interpreted as representing the official policies of the funding agencies or the government.

%% file: 8appendix.tex
\section{Notation} \label{ap: notation}
This paper uses a set of timestamped occurrences and edges to define the temporal graphs. Each node is associated with multiple timestamped edges at different timestamps. In Table~\ref{tab:temporal graph notations}, we list the main symbols in this paper to formalize the temporal graphs.
\begin{table}[h]
    \small
    \centering
    \captionsetup{skip=0pt}
    \caption{Notation explanation}
    \begin{tabular}{c|c}
        \hline
        Symbol & Description \\
        \hline
        $ \tilde{\mathcal{G}} = (\tilde{\mathcal{V}}, \tilde{\mathcal{X}}, \tilde{\mathcal{E}})$ & temporal graph \\
        \hline
        n & the total number of nodes \\
        \hline
        m & the total number of edges \\
        \hline
        $\mathcal{T}$ & the total number of timestamps in $\tilde{\mathcal{G}}$ \\
        \hline
        $ \tilde{\mathcal{V}} = \{ v_1^{1},...,v_{n}^{\mathcal{T}}\} $ & the set of temporal nodes in $\tilde{\mathcal{G}}$ \\
        \hline
        $ \tilde{\mathcal{E}} = \{ e_1^{1},...,e_{m}^{\mathcal{T}}\} $ & the set of temporal edges in $\tilde{\mathcal{G}}$ \\
        \hline
        $v = \{ v^{t_1},v^{t_2},... \}$ & node $v$ with its occurrences at ${t_1,t_2,...}$ \\
        \hline
        $e^t = (u^t, v^t)^t$ & temporal edge between $u^t$ and $v^t$ at $t$ \\
        \hline
        $\tilde{\mathcal{X}} = \{ \mathbf{x}_1^{1},...,\mathbf{x}_n^\mathcal{T} \}\in \mathcal{R}^{n \times d}$ & the set of features for each node in $\tilde{\mathcal{G}}$ \\
        \hline
        $\tilde{\mathbf{y}} = \{ \mathbf{y}_1^{1},...,\mathbf{y}_n^{\mathcal{T}} \}$ & the set of labels for each node in $\tilde{\mathcal{G}}$\\
        \hline
        $\tilde{\mathbf{y}} \in \{1, 2,..., K \}$ & the labels for nodes in $\tilde{\mathcal{G}}$ \\
        \hline
    \end{tabular}
    \label{tab:temporal graph notations}
\end{table}

\section{Theory Analysis} \label{ap: lemma proof}
In this section, we provide the proof for Lemma \ref{th: nonexchangeble CP}. We show the proof based on the idea that to satisfy the exchangeability condition, we need to assign different weights at different quantiles and then the coverage gap is bounded by weights and distribution difference.

\begin{proof}
Given a calibration set $\mathcal{D}_{c}$ and one random selected test data point from a $\mathcal{D}_{t}$, we can calculate the non-conformity score for the data points in the calibration set and the test data. The set of non-conformity score from $\mathcal{D}_{c}$ is denoted as $\Phi_{c}=\{s_1,\ldots,s_{n_{c}}\}$. The set of non-conformity score from $\mathcal{D}_{c}$ and the $j$th data point in $\mathcal{D}_{t}$ is denoted as $\Phi_j=\{s_1,\ldots,s_{n_{c}},s_{j_{t}}\}$, then we use $\Phi_j^i=\{(s_1,\ldots,s_{i-1},s_{j_{t}},s_{i+1},\ldots,s_{n_{c}},s_i)\}$ to denote the permutation of the test data point and the data points from calibration set.

According to the theory of conformal prediction, we know that if the non-conformity score of the time point does not satisfy the exchangeability condition, we can get the following function:
\begin{equation} \label{eq: exchangeability violation}
    y_{j_{t}} \not\in C_{j_{t}} \iff \ s_{j_{t}} > Q_{1-\alpha}\left(\sum_{i=1}^{n_{c} + 1} \tilde{\omega}_i \cdot
    \delta_{\Phi_{j}}\right),
\end{equation}
where $\delta$ means sorting the non-conformity in an ascending order. $\tilde{\omega}$ is the pre-defined weights with $\sum_i^{n_{c} + 1}\tilde{\omega}_i=1$ and under the exchangeability condition, $\tilde{\omega}_i=\frac{1}{n_{c} + 1}$. 

The Eq.~\ref{eq: exchangeability violation} shows the consequences when the exchangeability condition is violated. Thus, if we want the exchangeability condition to still hold, we want to prove that the quantile we get when calculating from other permutations of $\Phi_j^i$ should be no larger than the one we get from $\Phi_j$. Mathematically, we want to prove the following equations:
\begin{equation} \label{eq: target equation to be proved}
    Q_{1-\alpha}\left(\sum_{i=1}^{n_{c}} \tilde{\omega}_i \cdot
    \delta_{\Phi_{j}} + \tilde{\omega}_{n_{c}+1}\delta_{n_{c} + 1} \right) \geq Q_{1-\alpha}\left(\sum_{i=1}^{n_{c} + 1} \tilde{\omega}_i \cdot
    \delta_{\Phi_j^i}\right)
\end{equation}
From Eq.~\ref{eq: target equation to be proved}, we know that if the test point has the largest non-conformity score, then Eq.~\ref{eq: target equation to be proved} holds. We only have to prove whether the Eq.~\ref{eq: target equation to be proved} holds when the test point does not have the largest non-conformity score. 

Assuming that we have a permutation $\Phi_j^k$, then the quantile can be rewritten as 
\begin{equation} \label{eq: detailed quantile calculation with permutation}
    \sum_{i=1}^{k - 1}\tilde{\omega}_i s_{i} + \tilde{\omega}_k s_{n_{c}+1} + \sum_{i=k+1}^{n_{c}}\tilde{\omega}_i s_{i} + \tilde{\omega}_{n_{c}+1} s_{k},
\end{equation}
The quantile calculated from $\Phi_j$ can be rewritten as
\begin{equation} \label{eq: detailed quantile calculation without permutation}
    \sum_{i=1}^{k - 1}\tilde{\omega}_i s_{i} + \tilde{\omega}_k s_{k} + \sum_{i=k+1}^{n_{c}}\tilde{\omega}_i s_{i} + \tilde{\omega}_{n_{c}+1} s_{n_{c}+1},
\end{equation}

Recall that we want to show Eq.~\ref{eq: detailed quantile calculation without permutation} $\geq$ Eq.~\ref{eq: detailed quantile calculation with permutation} to prove that Eq.~\ref{eq: target equation to be proved} holds for any permutation $\Phi_j^k$. By subtracting Eq.~\ref{eq: detailed quantile calculation without permutation} and Eq.~\ref{eq: detailed quantile calculation with permutation}, we can get the following equation:
\begin{equation} \label{eq: subtraction between no permutation and permutation}
    (\tilde{\omega}_{n_{c} + 1} - \tilde{\omega}_k)(s_{n_{c} + 1} - s_k)
\end{equation}
To make Eq.~\ref{eq: subtraction between no permutation and permutation} $\geq 0$, we have to let $\tilde{\omega}_{n_{c} + 1} \geq \tilde{\omega}_k$ as exchangeability violation indicating that $s_{n_{c} + 1} > s_k$ and $s_k$ is the data point that leads to the violation of exchangeability condition. Then we know that 
$y_{j_{t}} \not\in C_{j_{t}} \implies \ s_k \in \Phi_j^k$
\begin{align}
    \mathbbm{P}(y_{j_{t}} \not\in C_{j_{t}}) &= \mathbbm{P}({s_k\in \phi_j^k} ) \nonumber
    =\sum_{i=1}^{n_{c}+1}\tilde{\omega}_i \cdot \mathbbm{P}({i \in \phi_j^k)}\\ \nonumber
    &\leq \sum_{i=1}^{n_{c}+1}\tilde{\omega}_i \cdot \left(\mathbbm{P}(i \in \Phi_j) + d_{TV}(\Phi_j,\Phi_j^k)\right)\\ \nonumber
    &=\mathbbm{E}[{\sum_{i \in \Phi_j} \tilde{\omega}_i}] +
    \sum_{i=1}^{n_{c}}\tilde{\omega}_i \cdot d_{TV}(\Phi_j,\Phi_j^k)\\ 
    \nonumber
    &\leq \alpha + \sum_{i=1}^{c}\tilde{\omega}_i\cdot d_{TV}(\Phi_j,\Phi_j^k), 
    \end{align}
\end{proof}

\section{Baselines and Datasets} \label{ap: baselines and datasets}
\subsection{Baselines Introduction} \label{ap: baselines introduction}
In this section, we provide a detailed description of each baseline. Particularly, TPS~\cite{Sadinle_2018}, APS~\cite{romano2020classification}, and RAPS~\cite{angelopoulos2020uncertainty} are crucial in conformal prediction, determining the target quantile value used to construct the prediction set. DAPS~\cite{zargarbashi2023conformal} is a non-conformity score specifically designed for static graphs as a competitive baseline to validate our approach's effectiveness further. CF-GNN~\cite{huang2023uncertainty} is a model-based approach that optimizes the APS non-conformity score by incorporating topological structures within a GNN framework. Then NEX\cite{barber2023conformal} is a non-exchangeability method that explicitly addresses the challenge of non-exchangeability in time series data. NAPS\cite{clarkson2023distribution} also assumes non-exchangeability on graphs and addresses this challenge by appropriately weighting the conformal scores to reflect the network structure in static graphs. The Unfolding GNN (UGNN) approach from \cite{davis2024valid} extends conformal prediction to temporal graphs while assuming exchangeability. However, UGNN was originally developed for static GNN architectures such as GCN, GAT, and GraphSAGE, which limits its direct applicability to temporal GNNs. To facilitate a fair comparison, we adapt UGNN by extracting node embeddings from the temporal GNN and applying a projection layer to generate final predictions.

\subsection{Datasets Introduction and Statistics} \label{ap: data statistics}
The \textit{WIKI dataset} consists of one month of edit history on Wikipedia pages, capturing the evolving nature of user interactions. The \textit{REDDIT dataset} includes one month of user-generated posts across various subreddits, with link features derived from text embeddings to represent user interactions. The \textit{DBLP dataset} is constructed from author profiles in the Digital Bibliography and Library Project (DBLP), providing insights into evolving scholarly networks. The \textit{IBM Anti-Money Laundering dataset} simulates financial interactions among individuals, companies, and banks, incorporating a subset of entities engaging in illicit activities to facilitate fraud detection research. Additionally, we analyze the temporal structures(Table~\ref{tab: statistics of datasets}, Table~\ref{tab: temporal statistics}) of these datasets to ensure diversity in both temporal dynamics and graph sizes, thereby covering a broad spectrum of real-world scenarios. The real-world data sets used in this paper, i.e., WIKI\footnote{\url{https://s3.us-west-2.amazonaws.com/dgl-data/dataset/tgl/WIKI}}, REDDIT\footnote{\url{https://s3.us-west-2.amazonaws.com/dgl-data/dataset/tgl/REDDIT}}, DBLP\footnote{\url{https://opendatalab.com/OpenDataLab/DBLP_Temporal}},
IBM\footnote{\url{https://github.com/IBM/AML-Data}} are publicly available and can be downloaded using the link we provide. 
\begin{table}[h]
    \centering
    \captionsetup{skip=0pt}
    \caption{Statistics of datasets}
    \begin{tabular}{ccccc}
    \hline
        Dataset & Category & Nodes & Edges & Time span \\ \hline
        WIKI & Social & 9227 & 157474 & 152757 \\
        REDDIT & Social & 10984 & 672447 & 669065 \\
        DBLP & Citation & 2390 & 146738 & 10 \\
        IBM & Financial & 515080 & 5078345 & 15018 \\ \hline
    \end{tabular}
    \label{tab: statistics of datasets}
\end{table}
\vspace{-10pt}
\section{Additional Results} \label{ap: additional experiments}
In this section, we provide more details about our experiments. We offer (i) accuracy tests for each best backbone temporal GNN; (ii) scalability tests to show \method's scalability; (iii) parameter analysis tests to show the influence of diffusion parameters; (iv) more analysis on the relation between temporal patterns and the performance of \method.
\begin{table}[h]
    \centering
    \captionsetup{skip=0pt}
    \caption{Accuracy for each backbone}
    \begin{tabular}{cccc}
        \hline
        \multirow{2}{*}{Datasets} & \multicolumn{3}{c}{Accuracy} \\ \cline{2-4}
        ~ & TGAT & JODIE & TGN \\ \hline
        WIKI & 0.901±0.026 & 0.955±0.034 & 0.962±0.045 \\ 
        REDDIT & 0.921±0.066 & 0.899±0.062 & 0.880±0.062 \\ 
        DBLP & 0.707±0.001 & 0.707±0.000 & 0.698±0.005 \\ 
        IBM & 0.924±0.034 & 0.869±0.039 & 0.921±0.029 \\ \hline
    \end{tabular}
    \label{tab: acc experiments}
\end{table}
\subsection{Accuracy test}
In this test, we provide test results using the accuracy metric for each backbone temporal GNN model and ensure that our backbone model reaches its best performance. Table~\ref{tab: acc experiments} summarizes the backbone model's accuracy on every dataset used in the paper.

\subsection{Temporal Patterns and Performance.} We provide statistics on the character of temporal patterns in each dataset and want to show how temporal patterns affect the performance of different models. We split all the time steps into ten intervals for each dataset and list the statistics in Table~\ref{tab: temporal statistics}, where we can see that these four datasets preserve different temporal patterns. The statistics of the number of nodes per interval demonstrate that the DBLP dataset has no node number change since interval 5, while others, like REDDIT and WIKI datasets, have both the nodes and edges development along all the time intervals. In the DBLP dataset whose nodes have fewer changes than others, \method achieves less significant improvement compared to other datasets. The reason, based on our understanding, is that \method focuses on capturing both the temporal and structural patterns of uncertainty in temporal graphs. Without much change in temporal dimension means less violation of the exchangeability condition which mitigates the effectiveness of our method. Another interesting finding is that the IBM dataset has fewer neighbors compared to other datasets. Under this condition, our method performs better as a node prediction value in a sparse graph tends to be easily affected by neighbors. A simple guess is that the propagation paths through the whole graph are sparse as well, which makes the influence of structural neighbors and temporal neighbors have more impact.
\begin{table}[ht!]
    \centering
    \small
    \setlength{\tabcolsep}{2pt}
    \captionsetup{skip=0pt}
    \caption{Statistics of temporal patterns for each dataset}
    \begin{tabular}{cccc|cccc}
    \hline
        \multicolumn{4}{c|}{WIKI} & \multicolumn{4}{c}{REDDIT} \\
        \hline
        Time & \# node & \# edge & \# neighbor & Time & \# node & \# edge & \# neighbor \\
        \hline
        1 & 2 & 1 & 1.0 & 1 & 2 & 1 & 1.0 \\
        2 & 2057 & 12356 & 8.2 & 2 & 8493 & 61208 & 8.1 \\
        3 & 3300 & 27158 & 10.4 & 3 & 9527 & 127548 & 14.9 \\
        4 & 4356 & 44063 & 12.3 & 4 & 10004 & 191783 & 21.2 \\
        5 & 5264 & 60718 & 13.7 & 5 & 10333 & 263056 & 28.1 \\
        6 & 6116 & 76952 & 14.7 & 6 & 10571 & 328746 & 34.3 \\
        7 & 6865 & 95245 & 16.0 & 7 & 10738 & 401360 & 41.1 \\
        8 & 7500 & 110946 & 16.9 & 8 & 10841 & 468326 & 47.5 \\
        9 & 8123 & 129100 & 18.1 & 9 & 10929 & 540258 & 54.3 \\
        10 & 8776 & 144881 & 18.6 & 10 & 10977 & 606947 & 60.7 \\
        \hline
        \hline
        \multicolumn{4}{c|}{DBLP} & \multicolumn{4}{c}{IBM} \\ \hline
        Time & \# node & \# edge & \# neighbor & Time & \# node & \# edge & \# neighbor \\
        \hline
        1 & 1602 & 2912 & 1.8 & 1 & 13440 & 10977 & 1.0\\
        2 & 1773 & 8320 & 4.7 & 2 & 493387 & 1698563 & 3.6\\
        3 & 1940 & 16170 & 8.3 & 3 & 511255 & 2194371 & 4.5\\
        4 & 2224 & 26818 & 12.1 & 4 & 512540 & 2923922 & 5.9\\
        5 & 2390 & 40408 & 16.9 & 5 & 513579 & 3779914 & 7.6\\
        6 & 2390 & 56650 & 23.7 & 6 & 514488 & 4767811 & 9.6\\
        7 & 2390 & 75570 & 31.6 & 7 & 515072 & 5077481 & 10.2\\
        8 & 2390 & 96926 & 40.6 & 8 & 515078 & 5077971 & 10.2\\
        9 & 2390 & 120778 & 50.5 & 9 & 515079 & 5078222 & 10.2\\
        10 & 2390 & 146738 & 61.4 & 10 & 515079 & 5078309 & 10.2\\ \hline        
    \end{tabular}
    \label{tab: temporal statistics}
\end{table}